\documentclass{article}

\usepackage[nonatbib,preprint]{neurips_2019}

\usepackage[dvipsnames]{xcolor}
\usepackage[utf8]{inputenc} % allow utf-8 input
\usepackage[T1]{fontenc}    % use 8-bit T1 fonts
\usepackage{hyperref}       % hyperlinks
\usepackage{url}            % simple URL typesetting
\usepackage{booktabs}       % professional-quality tables
\usepackage{amsfonts}       % blackboard math symbols
\usepackage{nicefrac}       % compact symbols for 1/2, etc.
\usepackage{microtype}      % microtypography

\usepackage{multirow}
\usepackage{makecell}
\usepackage{amssymb}
\usepackage{amsthm}
\usepackage{tikz}
\usepackage{pdfpages}
\usepackage{wrapfig, graphicx}
\usepackage{subcaption}

\usepackage{comment}
\usepackage{bbm}
\usepackage{color}
\usepackage{amsmath}
\DeclareMathOperator*{\argmax}{arg\,max}

\usepackage[linesnumbered,ruled]{algorithm2e}

\newtheorem{theorem}{Theorem}
\newtheorem{proposition}[theorem]{Proposition}

\newtheorem{remark}{Remark}

\newcommand\widebar[1]{\mathop{\overline{#1}}}
\title{Effective Medical Test Suggestions \\ Using Deep Reinforcement Learning}

\author{
Yang-En Chen \\ HTC Research \& Healthcare \\ \texttt{yangen\_chen@htc.com}
\And
Kai-Fu Tang \\ HTC Research \& Healthcare \\ \texttt{kevin\_tang@htc.com}
\And
Yu-Shao Peng \\ HTC Research \& Healthcare \\ \texttt{ys\_peng@htc.com}
\And
Edward Y. Chang \\ Stanford University \\ \texttt{eyuchang@gmail.com}
}

\begin{document}

\maketitle
\begin{abstract}
Effective medical test suggestions benefit both patients and physicians to conserve time and improve
diagnosis accuracy.
In this work, we show that an agent can learn to suggest effective medical tests.
We formulate the problem as a stage-wise Markov decision process and propose a reinforcement learning method to train the agent.
We introduce a new representation of multiple action policy along with the training method of the proposed representation.
Furthermore, a new exploration scheme is proposed to accelerate the learning of disease distributions.
Our experimental results demonstrate that the accuracy of disease diagnosis can be significantly improved with  good medical test suggestions.
\end{abstract}

\section{Introduction}

Artificial intelligence for medical test suggestions benefits our society in many ways.
For instance, an automated system for suggesting medical tests can save a patient's time.
A patient often spends a long time waiting to see a physician. Once the patient is able meet with the physician, they are then required to take medical tests and schedule another appointment to review their results and diagnosis. Artificial intelligence can expedite this process by enabling patients to take the system-suggested medical tests before their appointment with a doctor.
The system can suggest one or more medical tests.
If the system only suggests one medical test at a time, a patient may need to make multiple visits to the hospital to complete their tests and follow up with doctors.
A better approach is for a system to be capable of suggesting a set of medical tests at once to reduce the number of visits a patient must make to the hospital, which can be time consuming.

In this paper,
we consider the medical tests that generate numeric results (e.g., blood tests, urine tests, liver function testing, etc.) and do not include the radiologic tests, which involve medical images. 
To simulate the most common scenario in which doctors suggest medical tests,
the procedure of an automated system is designed to follow a sequential order.
The automated system first makes several symptom queries and then suggests multiple medical tests to disambiguate plausible hypotheses. Each symptom query should consider the answers to the previous queries to collect maximal information from a patient.  Once the automated system has exhausted productive queries, it suggests a set of medical tests that are expected to further improve diagnosis accuracy.

We consider the process of querying symptoms and suggesting medical tests to achieve a precise diagnosis to be an active feature acquisition problem.
Many algorithms for active feature acquisition have been proposed.
For instance, Bilgic and Getoor~\cite{DBLP:VOILA} proposed a graphical model which considers non-myopic feature acquisition strategies. Kanani and Melville~\cite{PTAFA} addressed
the feature acquisition problem for induction time.
Xu et al.~\cite{DBLP:CSTC13}, Kusner~\cite{DBLP:ASTC14}, and Nan and Saligrama~\cite{DBLP:Gbrt17} proposed decision tree related methods to select key features.
Recently, Janisch et al.~\cite{DBLP:ddqn} proposed a reinforcement learning (RL) approach to learn a query strategy. Their experimental results showed that the RL methods outperform other non-RL methods.

We formulate the medical test suggestion problem as a stage-wise Markov decision process and propose a reinforcement learning method to learn a strategy that can suggest multiple medical tests at a time.
In  traditional reinforcement learning,  the agent only performs an action at a time step so it cannot learn to suggest multiple medical tests at once.
We propose a representation for {\em multiple action policy} and a method for its training. Moreover, reinforcement learning methods are inefficient especially when the action space is large. We propose the {\em label-guided exploration}  to speed up the learning process.
We can access correct disease labels during the training process, so we use the label information to design an efficient learning algorithm.
The agent can suggest key medical tests to facilitate disease diagnosis when both these enhancements are used.
The experimental results indicate that our agent suggests an average of $4.77$ medical tests and achieves $89.75\%$ top-$5$ accuracy in disease diagnosis in the case of $200$ diseases.

\section{Stage-Wise Markov Decision Processes}

Here, we formally define our sequential decision making problem as a finite-horizon, discounted Markov Decision Process (MDP). 
An MDP $\mathcal{M}$ is a five tuple $(\mathcal{S},\mathcal{A},p,r,\gamma)$, where set $\mathcal{S}$ is the state space, set $\mathcal{A}$ is the action space, function $p$ is the transition mechanism, function $r$ is the reward distribution, and scalar $\gamma$ is the discount factor. 
Denote the agent's policy by $\pi$. Given the state $s$, the agent selects an action $a$ with the probability $\pi(s, a)$. 
When the agent performs an action at state $s_t$ at time step $t$, it receives reward $r_t$ generated by the MDP and transitions to the next state $s_{t+1}$. Assuming an episode terminates at time step $T$, we define the return of the agent at time step $t$ as $R_t = \sum_{i=t}^{T} \gamma^{i-t} r_i$. The objective of the agent is to find a policy that maximizes the expected return.

\paragraph{Stages.}
In this work, we partition an episode of an MDP into stages, which forms the {\em stage-wise MDP}.
The stage-wise MDP possesses two properties.
First, the agent can only choose actions from a specific action set in a given stage.
With this constraint, the agent focuses on a subset of actions in each stage.
Therefore, the action space is reduced, leading to more efficient learning.
Second, the agent is limited to following a sequential order to traverse different stages.
Since this limitation significantly reduces the possible sequences of actions, the search space becomes smaller. Therefore, the agent can explore a better sequence of actions with higher probability.
In this work, an episode is partitioned into three stages: the {\em symptom query stage} $e_\text{sym}$, the {\em medical test suggestion stage} $e_\text{med}$, and the {\em disease prediction stage} $e_\text{dis}$.
We detail our stage-wise Markov decision process in the following paragraphs.

\paragraph{States.}
Denote a patient by $x$, which is a random variable sampled from all patients.
At time step $t$, the state $s$ is a four tuple $s = (\mathcal{F}_t, x, t, e)$. 
The set $\mathcal{F}_t$ is the information obtainable at the state. The agent can only access the values of the information listed in $\mathcal{F}_t$.
We record the time step $t$ in the state, and the stage flag $e \in \{e_\text{sym}, e_\text{med}, e_\text{dis}\}$ indicates the state's stage.

In this work, the set $\mathcal{F}$ of information items  considers the demographic information items $\mathcal{T}$, the symptoms $\mathcal{U}$, and the medical test results $\mathcal{V}$.
For a patient $x$ and an information item $f \in \mathcal{T} \cup \mathcal{U} \cup \mathcal{V}$, we use the notation $f(x)$ to refer to the value of $f$ on the patient $x$. % \in \mathcal{T} \cup \mathcal{U} \cup \mathcal{V}$ of the patient $x$.
We say that a symptom $f \in \mathcal{U}$ is {\em present} on a patient $x$ if $f(x) = 1$ and is {\em absent} if $f(x) = -1$.
We also discretize the abnormal results of medical tests into categories and assign a positive integer to each category.
Therefore, we say that a medical test $f \in \mathcal{V}$ obtains a {\em normal} result on a patient $x$ if $f(x) = -1$ and gets an {\em abnormal} result if $f(x)$ is a positive integer.

\paragraph{Initial States.}
Every episode starts with an initial state satisfying the following conditions: (1) all demographic information $\mathcal{T}$ is known; (2) one of the present symptoms of the patient is given; (3) all the other information of the patient is unknown; (4) the state is in the symptom query stage. 
Formally, an initial state $s_0 = (\mathcal{F}_0, x, t, e)$ satisfies
\begin{equation*}
t=0\; \; \textrm{and} \; \;e = e_\text{sym}\; \; \textrm{and} \; \;\exists f \in \mathcal{U}\ldotp [\mathcal{F}_0 = \mathcal{T} \cup \{f\}\; \;\textrm{and} \; \;f(x) = 1 ].
\end{equation*}

\paragraph{Actions.} 
The actions of our problem can be categorized into three types: querying a symptom, suggesting a medical test, and predicting a disease, denoted by $\mathcal{A}_{\textrm{sym}}$, $\mathcal{A}_{\textrm{med}}$, and $\mathcal{A}_{\textrm{dis}}$, respectively. 
These three action sets represent the action spaces of the three stages $e_\text{sym}$, $e_\text{med}$, and $e_\text{dis}$.
Every symptom query in the action set $\mathcal{A}_{\textrm{sym}}$ targets a different symptom, and each action in $\mathcal{A}_{\textrm{dis}}$ predicts a different disease.
The actions in $\mathcal{A}_{\textrm{med}}$ correspond to all medical tests in $\mathcal{V}$.
Note that the agent shall suggests multiple medical tests at once, so it  selects a subset of actions $\Lambda \subseteq \mathcal{A}_{\textrm{med}}$ to perform when suggesting medical tests.

We add two actions $q_1$ and $q_2$ into $\mathcal{A}_{\textrm{sym}}$ in the symptom query stage so the agent can manage the timing to {\em quit} the stage. The action $q_1$ takes the agent to the medical test suggestion stage, and the action $q_2$ allows the agent to bypass test suggestions and enter the disease prediction stage.

In the symptom query stage, the agent can query a symptom or quit the stage. At a state $s_t = (\mathcal{F}_t, x, t, e_\text{sym})$, if the agent queries a symptom $f$, the transition to the next state $s_{t+1}$ is defined as
\begin{equation*}
\mathcal{F}_{t+1} = \mathcal{F}_t \cup \{f\},
 \; \;
s_{t+1} = (\mathcal{F}_{t+1}, x, t+1, e_\text{sym}).
\end{equation*}
When the agent performs the action $q_1$ or $q_2$ at the state $s_t = (\mathcal{F}, x, t, e_\text{sym})$ to quit the symptom query stage, the next state will be $(\mathcal{F}, x, t+1, e_\text{med})$ or $(\mathcal{F}, x, t+1, e_\text{dis})$. Note that $\mathcal{F}$ does not  gain new information through the transition, and the stage flag of the state is set to $e_\text{med}$ or $e_\text{dis}$ to indicate that the state has transitioned to the new stage.

In the medical test suggestion stage, the agent must select a set of medical tests from the power set $2^\mathcal{V}$. After the agent suggests a set of medical tests $\mathcal{W} \subseteq \mathcal{V}$ at a state $s_t = (\mathcal{F}_t, x, t, e_\text{med})$, the agent obtains the results of the suggested tests
and enters the disease prediction stage.
The state transition induced by the test suggestion is defined as
\begin{equation*}
\mathcal{F}_{t+1} = \mathcal{F}_t \cup \mathcal{W},
 \; \;
s_{t+1} = (\mathcal{F}_{t+1}, x, t+1, e_\text{dis}).
\end{equation*}

The agent is required to predict the patient's disease in the disease prediction stage $e_\text{dis}$.
When the agent predicts of the patient's disease, the episode ends with a reward or a penalty. 
The agent arrives at an artificial terminal state whenever the episode terminates.

\paragraph{Rewards.} 
We define the reward functions of the disease prediction stage, medical test suggestion stage, and symptom query stage in the following paragraphs.
First, in the disease prediction stage, the reward function consists of two components, {\em prediction reward} and {\em abnormality reward}.
In order to suggest medical tests that can maximize the disease-prediction accuracy, we design the {\em prediction reward} to encourage the agent to predict correctly and punish it for predicting wrong. Assuming the patient suffers from the disease $y$, and $\hat{y}$ is the agent's prediction, we define the prediction reward as 
\begin{equation*}
r_{\textrm{pred}}(y,\hat{y}) = 
\begin{cases}
    m & \text{if } y = \hat{y}, \\
    -n & \text{otherwise},
\end{cases}
\text{ where } m,n > 0.
\end{equation*}
The {\em abnormality reward} $r_{\text{ab}}$ is designed to encourage the agent to explore present symptoms and abnormal test results. Since a disease is usually reflected by symptoms and abnormal test results, these abnormalities provide key information for disease diagnosis. Given a state $s = (\mathcal{F},x,t,e)$,
we can compute the abnormality reward $r_{\text{ab}}(s)$ by adding {\em the number of known present symptoms} and {\em the number of abnormal medical test results}:
\begin{equation*}
    r_{\text{ab}}(s) = \lambda \times \left|\{f \mid f \in \mathcal{F} \cap ( \mathcal{U} \cup \mathcal{V})  \text{ and } f(x) > 0\}\right|,
\end{equation*}
where $\lambda$ is the weight of the rewards.
In the disease prediction stage, given  the agent's disease prediction $a$ for a patient who suffers from disease $y$, our reward function in the disease prediction stage is
\begin{equation*}
    r(s,a) = r_{\text{pred}}(y,a) + r_{\text{ab}}(s).
\end{equation*}

Second, the reward function in the medical test suggestion stage consists of the {\em medical test cost}. It attempt to balance the cost and effect of medical tests. In practice, given a disease, only particular medical tests are critical; therefore, a good strategy is to suggest a compact set of medical tests which is key to the diagnosis.
Therefore, when the agent takes the actions $\Lambda \subseteq \mathcal{A}_\text{med}$ to suggests a set of medical tests at a state $s$, we define the reward in the medical test suggestion stage as
\begin{equation*}
    r(s,\Lambda) = - c \times |\Lambda|,
\end{equation*}
where $c > 0$ is the cost of each medical test.
Note that we set the cost of all tests to be the same to optimize patient benefits, not hospital profit.

Third, in the symptom query stage, when the number of symptom queries is more than a predefined number $k$, the agent is considered as failing to predict correctly. In this case, the training episode terminates, and the agent receives a penalty of $-n$ and the abnormality reward $r_{\text{ab}}(s)$, where the former  agrees with the prediction reward and the latter  guides the agent to explore key information.
Therefore, when the agent selects an action $a \in \mathcal{A}_{\textrm{sym}}$ at a state $s = (\mathcal{F},x,t,e_\text{sym})$, the reward in the disease prediction stage is defined as
\begin{equation*}
    r(s,a) = 
\begin{cases}
    -n + r_{\text{ab}}(s) & \text{if } t \ge k \text{ and } a \not\in \{q_1, q_2\},
    \\
    0 & \text{otherwise}.
\end{cases}
\end{equation*}

\section{Methodology}

\subsection{Neural Network Model of the Agent}

\begin{wrapfigure}[18]{r}{6cm}
\centering
\begin{tikzpicture}[scale=0.12]

	\path (22, 20) coordinate(center);
	\path (50, 42) coordinate(bound);
	\clip (0,0) rectangle (bound);

	\path (0,0) -- (1,0)
	coordinate[pos=1] (w)
	coordinate[pos=0.5] (ww)
	coordinate[pos=-1] (nw)
	coordinate[pos=-0.5] (nww)
	;

	\path (0,0) -- (0,1)
	coordinate[pos=1] (h)
	coordinate[pos=0.5] (hh)
	coordinate[pos=-1] (nh)
	coordinate[pos=-0.5] (nhh)
	;

	\path (0,0) -- (0,7)
	coordinate[pos=1] (H)
	coordinate[pos=-1] (nH)
	;

	\path (0,0) -- (11,0)
	coordinate[pos=1] (W)
	coordinate[pos=-1] (nW)
	coordinate[pos=0.5] (WW)
	coordinate[pos=-0.5] (nWW)
	;

	\path (0,0) -- (8,2.5)
	coordinate[pos=1] (D)
	coordinate[pos=0.5] (DD)
	coordinate[pos=-1] (nD)
	coordinate[pos=-0.5] (nDD)
	;

	\path
	(w)+(h) coordinate(hw)
	(h)+(H) coordinate(hH)
	(nh)+(nH) coordinate(nhH)
	;
	
    \path
      (center) coordinate(e2)
      ++(nhH) coordinate(e1)
      ++(nhH) coordinate(e0)
      (e2)++(nWW)++(nW)
      ++(hH) coordinate(e31)
      ++(hH) coordinate(e41)
      (e2)++(nWW)
      ++(hH) coordinate(e32)
      ++(hH) coordinate(e42)
      (e2)++(WW)
      ++(hH) coordinate(e33)
      ++(hH) coordinate(e43)
      (e2)++(WW)++(W)
      ++(hH) coordinate(e34)
      ++(hH) coordinate(e44)
      ;

    \def\setmybox (#1) len (#2){
	    % vector box
	    \path (0,0) -- (DD)
	    coordinate[pos=#2] (d)
	    coordinate[pos=-#2] (nd)
	    ;
	    % vector box
	    \path
	    (nd) coordinate(fdl)
	    (nd)++(h) coordinate(ful)
	    (nd)++(w) coordinate(fdr)
	    (nd)++(hw) coordinate(fur)
	    (d) coordinate(bdl)
	    (d)++(w) coordinate(bdr)
	    (d)++(h) coordinate(bul)
	    (d)++(hw) coordinate(bur)
	    ;
        \path
        (#1)+(fdl) coordinate(#1_fdl)
        (#1)+(ful) coordinate(#1_ful)
        (#1)+(fdr) coordinate(#1_fdr)
        (#1)+(fur) coordinate(#1_fur)
        (#1)+(bdl) coordinate(#1_bdl)
        (#1)+(bdr) coordinate(#1_bdr)
        (#1)+(bul) coordinate(#1_bul)
        (#1)+(bur) coordinate(#1_bur)
        ;
        \path
        (#1_ful) -- (#1_fur) coordinate[pos=0.5] (#1_fu)
        (#1_fdl) -- (#1_fdr) coordinate[pos=0.5] (#1_fd)
        (#1_bul) -- (#1_bur) coordinate[pos=0.5] (#1_bu)
        (#1_bdl) -- (#1_bdr) coordinate[pos=0.5] (#1_bd)
        ;
    };

    \def\drawmybox (#1)[#2]{
        \draw[fill=#2, line join=round]
        (#1_fur) -- (#1_fdr) -- (#1_fdl) -- (#1_ful) -- cycle
        (#1_fur) -- (#1_ful) -- (#1_bul) -- (#1_bur) -- cycle
        (#1_fur) -- (#1_bur) -- (#1_bdr) -- (#1_fdr) -- cycle;
    };
    
    \def\definegriddepth (#1){
	    \path (0,0) -- (DD)
	    coordinate[pos=-#1] (gd)
	    ;
    };

    \def\drawboxgrid (#1)[#2]{
	    \path (0,0) -- (gd)
	    coordinate[pos=#2] (thisgd)
	    ;
        \draw[line join=round]
        (#1_bul)+(thisgd) coordinate(gbul)
        (#1_bur)+(thisgd) coordinate(gbur)
        (#1_bdr)+(thisgd) coordinate(gbdr)
        (gbul) -- (gbur) -- (gbdr);
    };

	% define box points
    \setmybox (e0) len (1);
    \setmybox (e1) len (1);
    \setmybox (e2) len (1);
    \setmybox (e31) len (1);
    \setmybox (e32) len (1);
    \setmybox (e33) len (1);
    \setmybox (e34) len (1);
    \setmybox (e41) len (1);
    \setmybox (e42) len (1);
    \setmybox (e43) len (1);
    \setmybox (e44) len (1);

	% network connect - right
    \draw (e0_bu) -- (e1_bd);
    \draw (e1_bu) -- (e2_bd);
    \draw  (e2_bu) -- (e31_bd);
    \draw (e31_bu) -- (e41_bd);
    \draw  (e2_bu) -- (e32_bd);
    \draw (e32_bu) -- (e42_bd);
    \draw  (e2_bu) -- (e33_bd);
    \draw (e33_bu) -- (e43_bd);
    \draw  (e2_bu) -- (e34_bd);
    \draw (e34_bu) -- (e44_bd);

	%\definecolor{head1}{HTML}{0444BF};
	%\definecolor{head2}{HTML}{058FF2};
	%\definecolor{head3}{HTML}{99DDFF};
	%\definecolor{head4}{HTML}{99F7DD};
	%\definecolor{encoder}{HTML}{FFFFFF};
 
    %\definecolor{head1}{HTML}{fbb034};
	%\definecolor{head2}{HTML}{ffdd00};
	%\definecolor{head3}{HTML}{c1d82f};
	%\definecolor{head4}{HTML}{00a4e4};
	%\definecolor{encoder}{HTML}{6a737b};
	\colorlet{head1}{Goldenrod}
	\colorlet{head2}{Orange}
	\colorlet{head3}{Bittersweet}
	\colorlet{head4}{RoyalBlue}
	\colorlet{encoder}{White}

	% draw box
    \drawmybox (e0)[encoder];
    \drawmybox (e1)[encoder];
    \drawmybox (e2)[encoder];
    \drawmybox (e31)[head1];
    \drawmybox (e32)[head2];
    \drawmybox (e33)[head3];
    \drawmybox (e34)[head4];
    \drawmybox (e41)[head1];
    \drawmybox (e42)[head2];
    \drawmybox (e43)[head3];
    \drawmybox (e44)[head4];
    \definegriddepth (0.26);
    \drawboxgrid (e41)[1];
    \drawboxgrid (e41)[2];

	% network connect - left
    \draw (e0_fu) -- (e1_fd);
    \draw (e1_fu) -- (e2_fd);
    \draw  (e2_fu) -- (e31_fd);
    \draw (e31_fu) -- (e41_fd);
    \draw  (e2_fu) -- (e32_fd);
    \draw (e32_fu) -- (e42_fd);
    \draw  (e2_fu) -- (e33_fd);
    \draw (e33_fu) -- (e43_fd);
    \draw  (e2_fu) -- (e34_fd);
    \draw (e34_fu) -- (e44_fd);

	% text
	\path (h)++(h) coordinate(output_text);
	\path (DD)++(hh)++(w) coordinate(layer_text);
	\path (w) coordinate(input_text);

	\draw (e0)++(input_text) node[anchor=north,align=center] {\footnotesize $s$};

	\def\layeroutput(#1)(#2){
		\draw (#1)++(output_text) node[anchor=south,align=center] {\scriptsize #2};
	};
    \layeroutput(e41)($\pi_\text{sym}$);
    \layeroutput(e42)($\pi_\text{med}$);
    \layeroutput(e43)($\pi_\text{dis}$);
    \layeroutput(e44)($\mathbf{z}$);
		
	\def\layersize(#1)(#2){
		\draw (#1)++(layer_text) node[anchor=west,align=left] {\tiny #2};
	};
    \layersize(e1)($2048$);
    \layersize(e2)($1024$);
    \layersize(e34)($1024$);
			
\end{tikzpicture}

    \caption{Multi-task neural network.}
    \label{fig:network}
\end{wrapfigure}
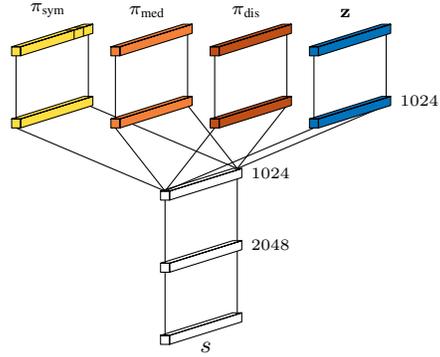

We use a multi-task neural network to model the agent. 
The input of the neural network considers the patient's information at a state $s$.
We encode the patient's information at the state as a {\em vector} in which each element corresponds to a symptom, a medical test result, or a demographic information item.
Consider a state $s = (\mathcal{F}, x, t, e)$ and an information item $f \in \mathcal{V} \cup \mathcal{U}$ which is a symptom or a medical test result. If $f$ is obtainable at the state $s$, i.e. $f \in \mathcal{F}$, the vector's element corresponding to $f$ is set to $f(x)$. Otherwise, the corresponding element is $0$ to indicate that $f$ is unknown.
On the other hand, we use the one-hot representation to discretize the demographic information.

Given the patient's information at a state, the network generates three distributions $\pi_\text{sym}$, $\pi_\text{med}$, and $\pi_\text{dis}$ to construct the agent's policy.
These three distributions are  defined over the action sets $\mathcal{A}_{\textrm{sym}}$, $\mathcal{A}_{\textrm{med}}$, and $\mathcal{A}_{\textrm{dis}}$ respectively.
The network also outputs a vector $\mathbf{z}$, which is used in an auxiliary task described in Section \ref{training_of_network}.

The multi-task neural network comprises a shared input encoder and four decoders.
Every encoder and decoder is composed of two fully-connected layers.
As depicted in Figure \ref{fig:network},
a state $s$ is encoded into the hidden representation through the encoder, and the hidden representation is then decoded into $\pi_\text{sym}$, $\pi_\text{med}$, $\pi_\text{dis}$, and $\mathbf{z}$ by their corresponding decoders.
A rectifier non-linearity is used in all layers except the output layers.
We use the softmax function $\phi(\mathbf{x}) = \frac{e^{\mathbf{x}}}{\sum_{j=1}^{n} e^{x_j}}$ in the decoders of $\pi_\text{sym}$ and $\pi_\text{dis}$.
Note that we use the sigmoid function $\sigma(x) = \frac{1}{1 + e^{-x}}$ in the decoder of $\pi_\text{med}$, which will be explained later.
The sigmoid function is also used in the decoder of $\mathbf{z}$ because the auxiliary task aims to predict a binary vector.

Although the network generates three distributions at each state, only the distribution over the valid action set in the stage forms the agent's policy. Specifically, given a state $s = (\mathcal{F}, x, t, e)$, 
we define the agent's policy at the state $s$ as
\begin{equation*}
\pi(s,a) = 
\begin{cases}
    \pi_{\text{sym}}(s,a) & \text{if } e = e_\text{sym}, \\
    \pi_{\text{med}}(s,a) & \text{if } e = e_\text{med}, \\
    \pi_{\text{dis}}(s,a) & \text{if } e = e_\text{dis}.
\end{cases}
\end{equation*}

\paragraph{Multiple Action Policy.} In the typical RL setting, a {\em single} action is performed at each step. To suggest a set of medical tests for a patient, we would like to perform {\em multiple} actions at once. A naive method is to expand all combinatorial actions in $\mathcal{A}_{\textrm{med}}$ to the action space. This results in $|2^{\mathcal{A}_{\textrm{med}}} |$ actions, and the number of parameters blow up exponentially in terms of the size of the original action space $\mathcal{A}_{\textrm{med}}$.
To achieve efficient multiple action selection, we need a compact representation of $\pi_{\textrm{med}}$ such that the parameters of $\pi_{\textrm{med}}$ scales up {\em linearly} in terms of the size of $\mathcal{A}_{\textrm{med}}$.
We propose a new representation of {\em multiple action policy}  to solve this difficulty.

In the traditional setting, the softmax function is used in the output layer of $\pi_{\textrm{med}}$; in the multiple action setting, we partition the policy $\pi_{\textrm{med}}$ into $\pi_a$, for each $a\in\mathcal{A}_{\textrm{med}}$. We use the sigmoid function in the output layer of $\pi_a$. Therefore, the output value of $\pi_a$ can be interpreted as the probability of selecting an action $a \in \mathcal{A}_{\textrm{med}}$. Using this new representation, the probability of selecting multiple actions $\Lambda\in 2^{\mathcal{A}_{\textrm{med}}}$ in a state $s$ can be calculated by
\begin{equation}
\pi_{\textrm{med}}(s, \Lambda) = \prod_{a\in \Lambda} \pi_a(s) \prod_{a\not\in \Lambda} (1 - \pi_a(s)).\label{new_pi}
\end{equation}

\begin{proposition}
The multiple action policy $\pi_{\textrm{med}}$  defined by Equation~\ref{new_pi} satisfies
\begin{equation*}
\sum_{\Lambda\in 2^{\mathcal{A}_{\textrm{med}}}} \pi_{\textrm{med}}(s,\Lambda) = 1 \text{ for all } s.\label{new_pi_property}
\end{equation*}
\label{prop:1}
\end{proposition}

\begin{proposition}
The set $\Lambda$ of multiple actions  which achieves the maximum probability in $\pi_{\textrm{med}}$ can be characterized by
\begin{equation*}
\left\{ a \in  \mathcal{A}_{\textrm{med}}  \mid \pi_a(s) \ge \frac{1}{2} \right\} \in \argmax_{\Lambda\in 2^{\mathcal{A}_{\textrm{med}}}} \pi_{\textrm{med}}(s, \Lambda) .
\end{equation*}
\label{prop:2}
\end{proposition}

\begin{remark}
The multiple action policy $\pi_{\textrm{med}}$ is a legitimate probability distribution over action space $2^{\mathcal{A}_{\textrm{med}}}$. While the size of action space is exponential in $|\mathcal{A}_{\textrm{med}}|$, the parameters used in $\pi_{\textrm{med}}$ only scales up {\em linearly} in $|\mathcal{A}_{\textrm{med}}|$. More importantly, selecting the best multiple actions from $\pi_{\textrm{med}}$  can be performed in {\em linear} time.
\end{remark}
The proofs of Proposition~\ref{prop:1} and~\ref{prop:2} are available in the supplementary material.

\subsection{Training of the Neural Network}
\label{training_of_network}
We train our neural network with the policy-based  REINFORCE method~\cite{policy_gradient_2000}   to maximize the expected total return $\mathbb{E}_{\pi}[\sum_{t=0}^{T} \gamma^t r_t]$. We propose three techniques to improve  training efficiency and overall performance. We present these techniques first and then describe the objective function.

\paragraph{Multiple Action Sampling.}
Multiple action sampling can be performed using our new representation of policy $\pi_{\textrm{med}}$. To sample multiple actions from $\pi_{\textrm{med}}$,
we can  sample individual $x_a$ from Bernoulli distributions $\pi_a$, that is, 
$x_a \sim \pi_a$ for each  $a \in \mathcal{A}_{\textrm{med}}$.  %.\label{sample}
Then, by Equation~\ref{new_pi}, the multiple actions $\Lambda$ can be assembled from the individual sampled results $x_a$, 
i.e., 
$\Lambda = \{a \in \mathcal{A}_{\textrm{med}} \mid x_a = 1\}$.

The new representation of our multiple action policy needs to be updated by gradient descent during the REINFORCE algorithm. The traditional policy gradient is computed by
\begin{equation*}
\nabla_\theta \mathbb{E}_\pi[R] \propto \mathbb{E}_\pi \left[ \nabla_\theta  R  \log \pi(s,a)  \right].
\end{equation*}
Therefore, the policy gradient used in our new representation $\pi_{\textrm{med}}$ can be computed by
\begin{align*}
\nabla_\theta R \log \pi_{\textrm{med}}(s,\Lambda) &= \nabla_\theta R \log \prod_{a\in \Lambda} \pi_a(s) \prod_{a\not\in \Lambda} (1 - \pi_a(s)) \\
&= \sum_{a\in\Lambda} \nabla_\theta R \log \pi_a(s) + \sum_{a\not\in\Lambda} \nabla_\theta R \log (1 - \pi_a(s)).
\end{align*}

\paragraph{Entropy Regularizer.}
We use the entropy regularizer~\cite{DBLP:a3c,DBLP:entropy_reg} to prevent the agent from converging to sub-optimal policies.
We compute the entropy $H$ of the distributions $\pi_\text{sym}(s,\cdot)$ and $\pi_\text{dis}(s,\cdot)$ by
$H(\pi(s,\cdot)) = - \sum_{a \in \mathcal{A}} \pi(s,a) \log \pi(s,a)$.
To compute the entropy of $\pi_\text{med}$,
we can treat $\pi_a$'s  as independent distributions (Equation~\ref{new_pi}).
Therefore, by the property of the sum of individual entropies~\cite{DBLP:EoIM}, we have
\begin{equation*}
    H(\pi_\text{med}(s,\cdot))
    = - \sum_{a \in \mathcal{A}_\text{med}} H(\pi_a(s)).
\end{equation*}

\paragraph{Guiding Exploration with Disease Labels.}
We propose an exploration strategy called \emph{label-guided exploration} to speed up the learning process.
In our training process, the agent needs to learn the policy $\pi_{\text{sym}}$ of symptom queries, the policy $\pi_{\text{med}}$ of test suggestions, and the disease distribution $\pi_{\text{dis}}$. However, in the RL setting, learning the distribution $\pi_{\text{dis}}$ is inefficient. To learn the distribution $\pi_{\text{dis}}$, the agent needs to predict a disease and update $\pi_{\text{dis}}$ based on the reward.
Given that there are hundreds of possible diseases, the probability of exploring the correct label is quite low.
To remedy this problem,
we  can use  the correct labels, which are available during our training process, to guide the learning.
At the disease prediction stage, with a probability of $\epsilon$, we force the agent to choose the correct label instead of the label sampled from $\pi_{\text{dis}}$.
\begin{comment}
Assume that at a state $s = (\mathcal{F},x,2)$, the correct prediction for the patient $x$ is $a_x \in \mathcal{A}_{\text{dis}}$.
The agent selects the action $a$ with
\begin{equation*}
a = 
\begin{cases}
    a_x,                 & \text{if } \xi < \epsilon,
    \\
    a \sim \pi(s,\cdot), & \text{otherwise}.
\end{cases}
\end{equation*}
\end{comment}
This exploration scheme can accelerate policy learning.

Although label-guided exploration is helpful in learning,
a large $\epsilon$ may degrade training performance.
The training of $\pi_{\text{sym}}$ and $\pi_{\text{med}}$ aims to learn a policy that collects critical symptoms and test results in order to differentiate between diseases.
To achieve this, we need to evaluate the policy with the agent's predictions and update the policy accordingly.
A large $\epsilon$ will cause that most of the agent's predictions are ignored and replaced with the correct labels, leading to inaccurate policy evaluation. This consequently impedes the training of $\pi_{\text{sym}}$ and $\pi_{\text{med}}$.

The probability $\epsilon$ should be set to a reasonable value to train all the distributions $\pi_{\text{sym}}$, $\pi_{\text{med}}$, and $\pi_{\text{dis}}$. 
In one extreme case where $\epsilon$ is set to $1$, the training of $\pi_{\text{dis}}$ is efficient, but the agent cannot learn $\pi_{\text{sym}}$ and $\pi_{\text{med}}$.
In contrast, setting $\epsilon$ to too small a value diminishes the effect of label-guided exploration and provide little help to the learning of $\pi_{\text{dis}}$.
The probability $\epsilon$ of label-guided exploration balances the training efficiency of $\pi_{\text{dis}}$ and the performance of $\pi_{\text{sym}}$ and $\pi_{\text{med}}$.

\paragraph{Training.}
The network is trained on two tasks: optimizing the agent's policy and learning to rebuild the patient's full information.
We impose the auxiliary task of rebuilding  on the network to 
learn the correlation between symptoms and medical test results~\cite{DBLP:REFUEL}.
Denote all symptoms by $f_1, f_2, ..., f_{|\mathcal{V}|}$
and all medical tests by $f_{|\mathcal{V}|+1}, f_{|\mathcal{V}|+2}, ..., f_{|\mathcal{V}|+|\mathcal{U}|}$.
Given a patient $x$, we define the vector $\mathbf{g} \in \{0,1\}^{|\mathcal{V}|+|\mathcal{U}|}$ by
\begin{equation*}
    g_i = 
    \begin{cases}
        1 & \text{if } f_i(x) > 0, \\
        0 & \text{otherwise}.
    \end{cases}
\end{equation*}
The vector $\mathbf{g}$ marks the patient's present symptoms and abnormal test results.
Recall that the output vector $\mathbf{z}$  in Figure \ref{fig:network} is generated for this auxiliary task.
We train the network to rebuild $\mathbf{g}$ by the output vector $\mathbf{z}$.
Given the output $\mathbf{z}$ and the the vector $\mathbf{g}$, we define the rebuilding loss $l(\mathbf{z}, \mathbf{g})$ with the binary cross entropy:
\begin{equation*}
    l(\mathbf{z}, \mathbf{g}) = 
    -\mathbf{g}^{\intercal} \log \mathbf{z} - (\mathbf{1} - \mathbf{g})^{\intercal} \log (\mathbf{1} - \mathbf{z}).
\end{equation*}

Denote the parameters of our neural network by $\theta$.
We define the expected return $J_{\text{ret}}(\theta)$,
the entropy regularizer $J_{\text{ent}}(\theta)$, and
the rebuilding loss $J_{\text{reb}}(\theta)$:
\begin{equation*}
J_{\text{ret}}(\theta) = \mathbb{E}_{\pi_\theta}\left[
\sum_{t=0}^{T} \gamma^{t} r_t
\right],
\;\;
J_{\text{ent}}(\theta) = \mathbb{E}_{\pi_\theta} \left[
\sum_{t=0}^{T} H(\pi(s_t, \cdot))
\right],
\text{ and }
J_{\text{reb}}(\theta) = \mathbb{E}_{\pi_\theta} \left[
\sum_{t=0}^{T} l(\mathbf{z}_t, \mathbf{g})
\right].
\end{equation*}
Therefore, we have the objective function 
\begin{equation*}
J(\theta) = 
J_{\text{ret}}(\theta) +
\beta  J_{\textrm{ent}}(\theta) - 
\kappa  J_{\text{reb}}(\theta)
,
\end{equation*}
where $\beta$  and $\kappa$ are hyperparameters.
We update the network's parameters  with a learning rate $\alpha$ to maximize the objective function $J(\theta)$.

\section{Experiments}

\begin{wraptable}{r}{6cm}
	\centering	
	\caption{The hyperparameters selected by Hyperband.}
	\begin{tabular}{l|r}
    	\toprule
    	Hyperparameter  & Value\\
    	\midrule
    	Correct prediction reward $m$ & $0.8743$\\
    	Wrong prediction reward $n$ & $-0.7075$\\
        Medical test cost $c$ & $-0.0084$\\
        Abnormality reward factor $\lambda$ & $0.1915$\\
        Entropy regularizer $\beta$ & $0.0117$\\
        Exploration factor $\epsilon$ & $0.0056$\\
    	\bottomrule
	\end{tabular}
	\label{tab:hyper}
\end{wraptable}

To train an agent in our stage-wise sequential decision problem, we simulate the symptoms, medical test results, and correct disease labels of patients. We synthesize our data  according to a set of conditional probability tables from hospital.
These tables describe the conditional probability distributions of the symptoms, demographic information, and medical test results given a patient's disease. The disease labels of patients are sampled first. We uniformly sample the patients' diseases to avoid data imbalance.
Afterward, we sample the symptoms, demographic information, and medical test results of each patient from the conditional distributions  given the disease label.
We additionally sample a present symptom of each patient as the initial symptom which is given to the agent at the initial state.

All of our experiments are performed with the same settings.
We use a simulated dataset of $10^6$ patients for training and two different simulated datasets of $10^5$ patients for validation and testing.
We use Adam~\cite{DBLP:adam} as our optimizer. 
The learning rate $\alpha$ is $10^{-4}$ and the batch size is $512$.
The parameter for limiting the number of symptom queries $k$ is $9$.
The discount factor $\gamma$ is $0.99$.
The coefficient of the rebuilding loss $\kappa$ is $10$.
In additioin to these hyperparameters, we employ Hyperband~\cite{hyperband} to tune other critical hyperparameters including the prediction reward $m, n$, the cost $c$ for each medical test, the coefficient $\lambda$ of the abnormality reward, and the  $\epsilon$ for label-guided exploration.
The hyperparameters selected by Hyperband are reported in  Table~\ref{tab:hyper}.

\begin{figure}[ht]
    \begin{subfigure}{.33\linewidth}
        \centering
        \includegraphics[trim=130 480 200 100, clip, width=1.8in]{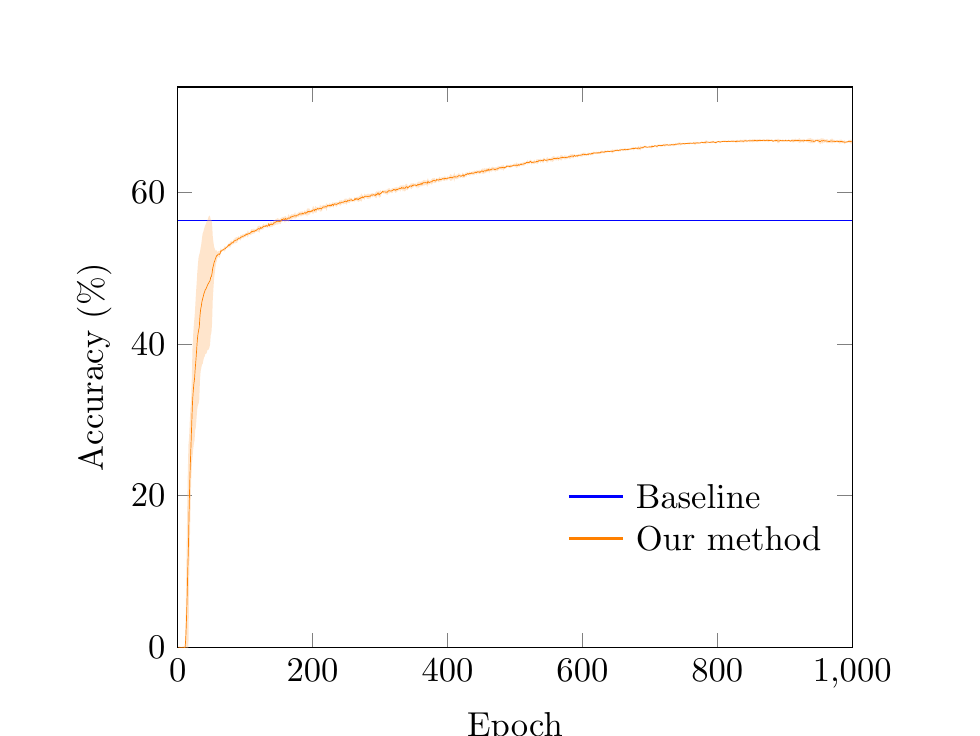}
        \caption{$200$ diseases}
    \end{subfigure}
    \begin{subfigure}{.33\linewidth}
        \centering
        \includegraphics[trim=130 480 200 100, clip, width=1.8in]{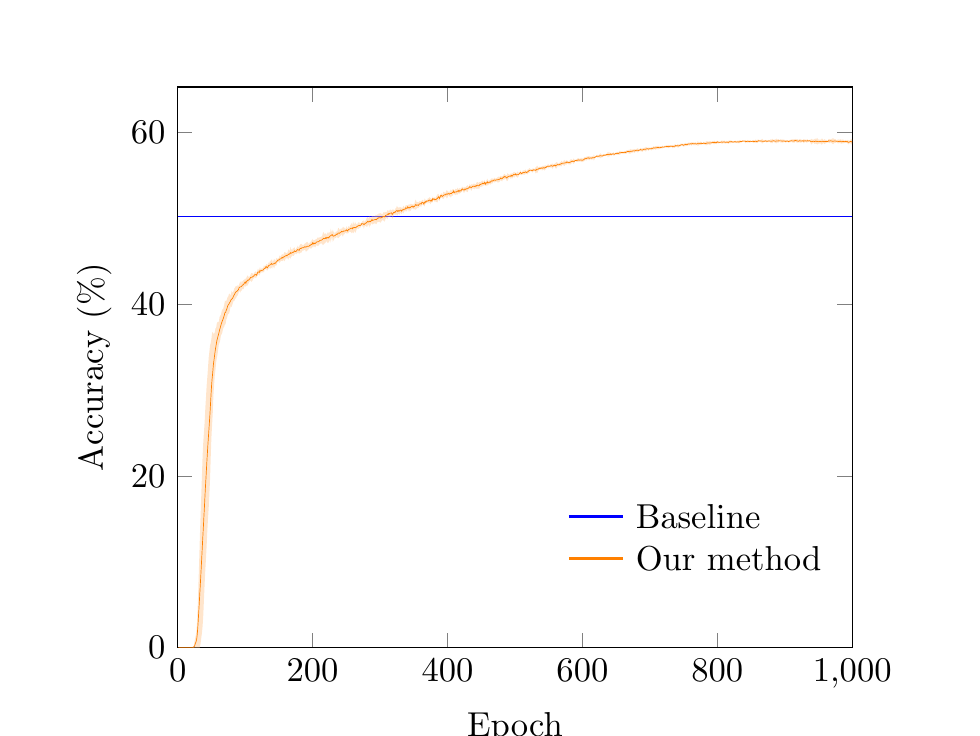}
        \caption{$300$ diseases}
    \end{subfigure}
    \begin{subfigure}{.33\linewidth}
        \centering
        \includegraphics[trim=130 480 200 100, clip, width=1.8in]{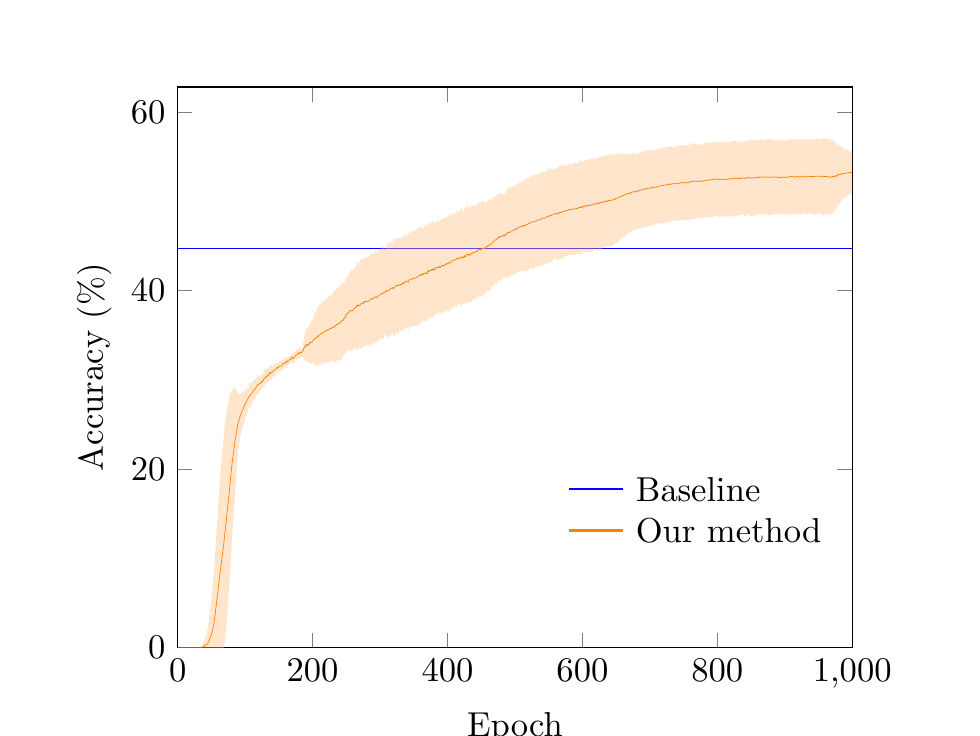}
        \caption{$400$ diseases}
    \end{subfigure}
    \caption{The training accuracy of our method (orange curve) and the baseline (blue line) on three different datasets.}
    \label{fig:training_acc}
\end{figure}

\begin{figure}[ht]
    \begin{subfigure}{.33\linewidth}
        \centering
        \includegraphics[trim=130 480 200 100, clip, width=1.8in]{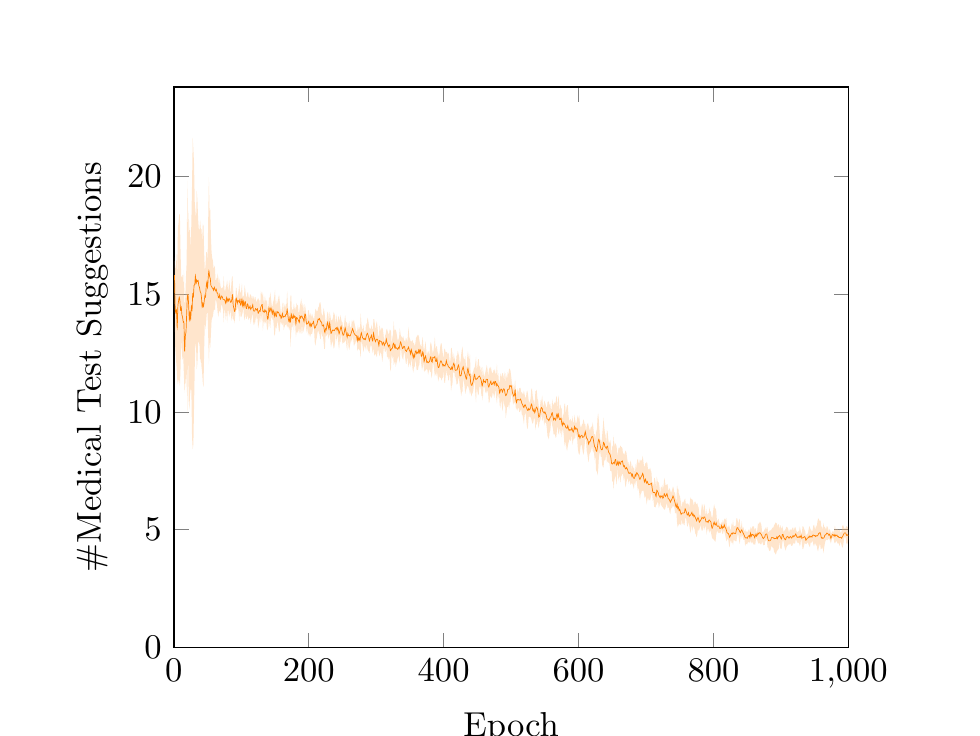}
        \caption{$200$ diseases}
    \end{subfigure}
    \begin{subfigure}{.33\linewidth}
        \centering
        \includegraphics[trim=130 480 200 100, clip, width=1.8in]{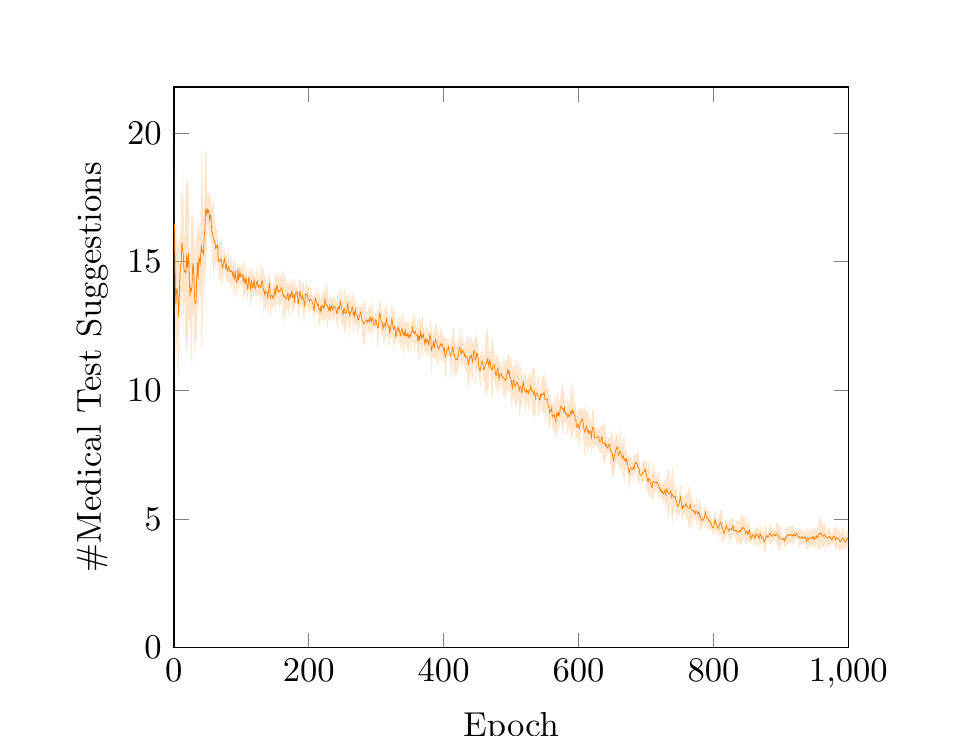}
        \caption{$300$ diseases}
    \end{subfigure}
    \begin{subfigure}{.33\linewidth}
        \centering
        \includegraphics[trim=130 480 200 100, clip, width=1.8in]{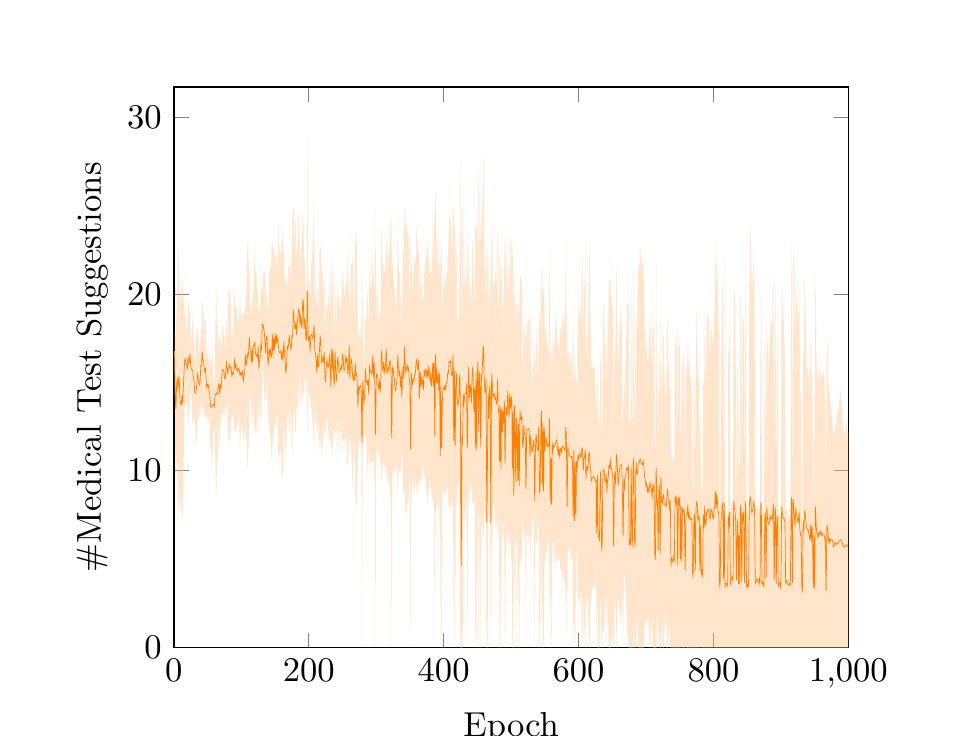}
        \caption{$400$ diseases}
    \end{subfigure}
    \caption{The average number of medical test suggestions.}
    \label{fig:num_test}
\end{figure}

\begin{figure}[ht]
    \begin{subfigure}{.33\linewidth}
        \centering
        \includegraphics[trim=130 480 200 100, clip, width=1.8in]{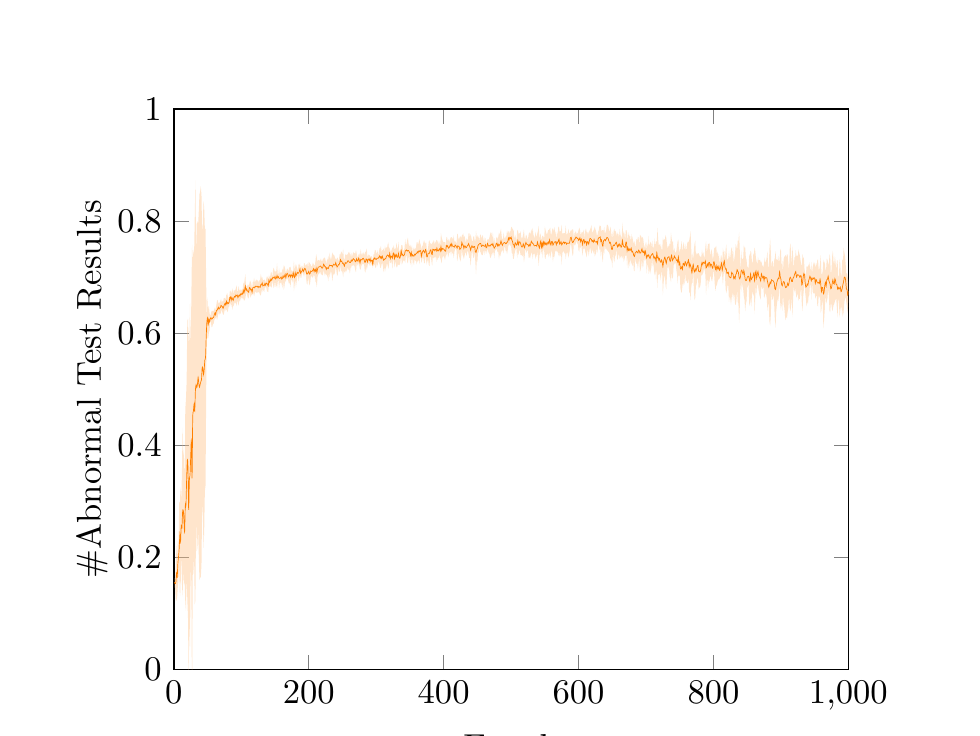}
        \caption{$200$ diseases}
    \end{subfigure}
    \begin{subfigure}{.33\linewidth}
        \centering
        \includegraphics[trim=130 480 200 100, clip, width=1.8in]{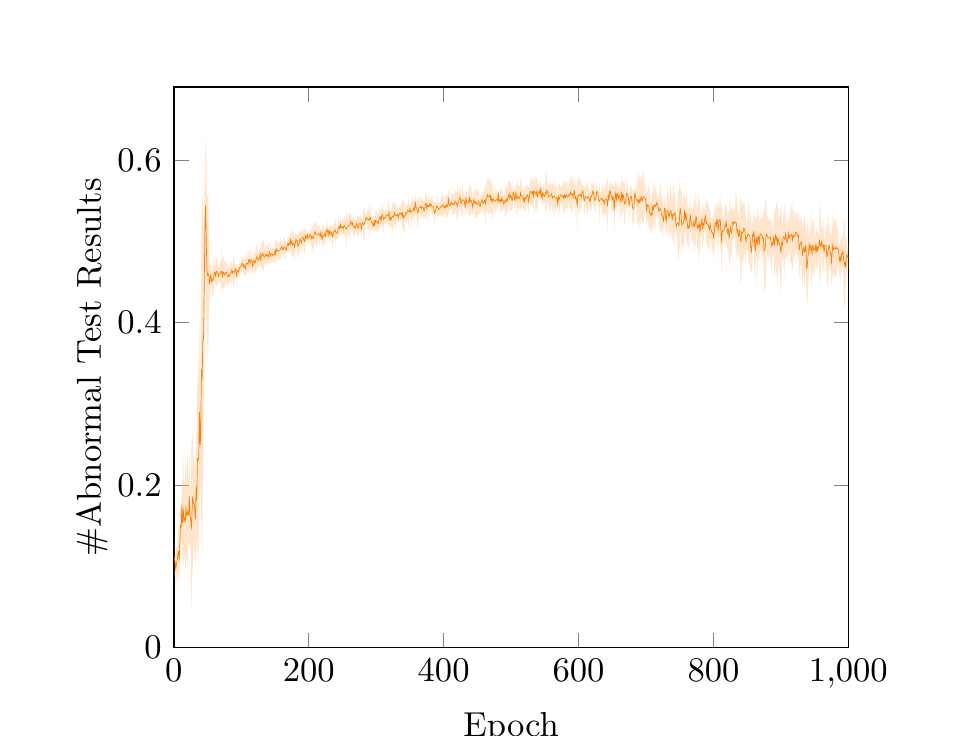}
        \caption{$300$ diseases}
    \end{subfigure}
    \begin{subfigure}{.33\linewidth}
        \centering
        \includegraphics[trim=130 480 200 100, clip, width=1.8in]{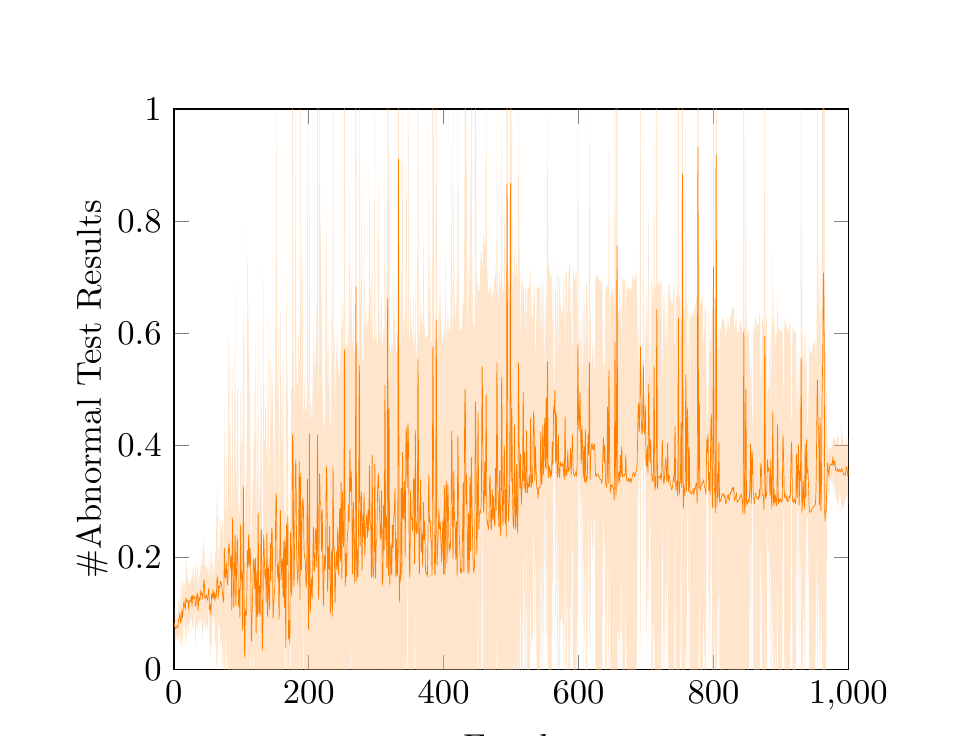}
        \caption{$400$ diseases}
    \end{subfigure}
    \caption{The average number of abnormal test results.}
    \label{fig:num_ab_test}
\end{figure}

To evaluate the extent to which a set of medical tests can help disease diagnosis, we perform three experiments with $200$, $300$, and $400$ diseases respectively, where the diseases are selected based on the disease frequency.
In these three experiments, we consider $36$ common medical tests.
We measure the improvement in diagnosis accuracy given the set of medical tests. The accuracy improvement is determined by the difference between two scenarios:
(1) disease prediction with symptom queries only, and (2) disease prediction with both symptom queries and medical test results.
We report the comparison of the training and test accuracy between the approaches where the medical test results are and are not given. We also show the average number of medical test suggestions provided by our approach.

Figure~\ref{fig:training_acc} shows the comparison of the disease diagnosis accuracy between the baseline and our proposed method during training.
We use REFUEL~\cite{DBLP:REFUEL} as the baseline algorithm, which is the state-of-the-art diagnosis model without considering medical test information.
In Figure~\ref{fig:training_acc}, the x-axis is the training epoch and the y-axis is the training accuracy.
The orange curve is the average training accuracy of our method over five different random seeds, and the shaded area represents two standard deviations. The blue line is the performance of the baseline algorithm.
We observe that the training accuracy of our proposed method (with medical test suggestion) outperforms the baseline by about $10\%$ in all cases.

Next, we investigate the agent's behavior of medical test suggestions during training.
Figure~\ref{fig:num_test} shows the average number of suggested medical tests provided by the agent.
We can see that the average number of suggested medical tests  decreases over time.
Figure~\ref{fig:num_ab_test} indicates the number of abnormal medical test results, which sharply increases at the beginning of the training process and then is sustained.
Considering both figures, we conclude that the agent can learn to avoid suggesting too many medical tests and suggest only critical ones. 

To measure the performance of our proposed method after training, we select the models based on the validation set and evaluate them on the testing set.
Table~\ref{tab:test_acc} shows the test accuracy.
The test accuracy 
reported in Table~\ref{tab:test_acc} are averaged over five different random seeds.
The test results indicate that our method outperforms the baseline in all cases.
Concretely, we observe a $7\%$-$14\%$ improvement when we consider medical test suggestions.

In Table~\ref{tab:test_num}, the suggestion ratio is the probability that the agent provides medical test suggestions for a patient; the number of suggested tests is the average number of the suggested tests of a patient; the abnormality discovery ratio is the ratio of the number of abnormal test results discovered by the agent to the total number of abnormal test results from  our dataset.
In our experiments, we consider $36$ medical tests.
The agent chooses to suggest medical tests with a probability around $35$-$48\%$ because not every patient needs medical tests.
When the agent suggests medical tests, it suggests $4$-$6$ medical tests on average.
The tests suggested by the agent can discover over $90\%$ of abnormalities in the cases of $200$ and $300$ diseases.
Therefore, we conclude that our agent can suggest critical medical tests.

\begin{table}
	\centering	
	\caption{The test accuracy of our method and baseline.}
	\begin{tabular}{c|rrr|rrr}
    	\toprule
    	\#Diseases  & \multicolumn{ 3}{c}{Baseline} & \multicolumn{ 3}{|c}{Our method}\\
    	& Top $1$ & Top $3$ & Top $5$ & Top $1$ & Top $3$ & Top $5$\\
    	\midrule
    	$200$ & $54.69$ & $74.63$ & $80.88$ & $\textbf{64.74}$ & $\textbf{83.74}$ & $\textbf{89.75}$\\
    	$300$ & $48.28$ & $67.31$ & $74.04$ & $\textbf{56.18}$ & $\textbf{77.05}$ & $\textbf{84.87}$\\
        $400$ & $42.82$ & $59.49$ & $65.60$ & $\textbf{49.83}$ & $\textbf{71.09}$ & $\textbf{79.73}$\\
    	\bottomrule
	\end{tabular}
	\label{tab:test_acc}
\end{table}

\begin{table}
	\centering	
	\caption{The statistics of medical test suggestions.}
	\begin{tabular}{c|ccc}
    	\toprule
    	\#Diseases  & Suggestion Ratio (\%) & \#Suggested Tests & Abnormality Discovery Ratio (\%)\\
    	\midrule
    	$200$ & $47.59$ & $4.77$ & $92.81$\\
    	$300$ & $43.40$ & $4.31$ & $91.22$\\
        $400$ & $35.60$ & $6.03$ & $73.70$\\
    	\bottomrule
	\end{tabular}
	\label{tab:test_num}
\end{table}

\section{Conclusions}
In this work, we demonstrated that an agent can learn to suggest medical tests to facilitate disease diagnosis.
We formulated the problem as a stage-wise Markov decision process and proposed a reinforcement learning method 
for training the agent.
We introduced a new multiple action policy representation along with the training method of the proposed representation.
Furthermore, a new exploration scheme was proposed to accelerate the learning of disease distributions.
Our experimental results showed that the accuracy of disease diagnosis can be significantly improved with medical tests.

%\bibliographystyle{abbrv}
%\bibliography{neurips_2019}

\newpage
\appendix
\section{Proofs of Propositions}

Recall that we define a multiple action policy over the action set $\mathcal{A}$ as
\begin{equation}
\label{new_pi_appendix}
\pi(s, \Lambda) = \prod_{a\in \Lambda} \pi_a(s) \prod_{a\in\mathcal{A}\setminus\Lambda} (1 - \pi_a(s)).
\end{equation}
Then, we have two propositions. 
\paragraph{Proposition~\ref{prop:1}.}
{\em
The multiple action policy $\pi$  defined by Equation~\ref{new_pi_appendix} satisfies
}
\begin{equation*}
\sum_{\Lambda\in 2^{\mathcal{A}}} \pi(s,\Lambda) = 1 \text{ for all } s.
\end{equation*}

\paragraph{Proposition~\ref{prop:2}.}
{\em
The  set $\Lambda$ of multiple actions  which achieves the maximum probability in $\pi$ can be characterized by
}
\begin{equation*}
\left\{ a \in  \mathcal{A}  \mid \pi_a(s) \ge \frac{1}{2} \right\} \in \argmax_{\Lambda\in 2^{\mathcal{A}}} \pi(s, \Lambda) .
\end{equation*}

Here we give the proofs of Proposition~\ref{prop:1} and~\ref{prop:2}.

\subsection{Proof of Proposition 1}
We prove this proposition by induction. Consider the case  $\mathcal{A} = \{b\}$. We have
\begin{align*}
\sum_{\Lambda\in 2^{\mathcal{A}}} \pi(s,\Lambda)
&= 
\sum_{\Lambda\in \{\emptyset,\{b\}\}} \pi(s,\Lambda)
\\&=
\prod_{a\in \emptyset} \pi_a(s) \prod_{a\in\{b\}} (1 - \pi_a(s))
+
\prod_{a\in \{b\}} \pi_a(s) \prod_{a\in \emptyset} (1 - \pi_a(s))
\\&= 1 - \pi_b(s) + \pi_b(s)
\\&= 1.
\end{align*}
Assuming that the statement holds given a set of actions $\mathcal{A}$,
we show that the statement holds on the set $\mathcal{A}' = \mathcal{A} \cup \{b\}$.
\begin{align*}
\sum_{\Lambda\in 2^{\mathcal{A}'}} \pi(s,\Lambda)
&=
\sum_{\Lambda\in 2^{\mathcal{A}'}}
\prod_{a\in \Lambda} \pi_a(s)
\prod_{a\in \mathcal{A}'\setminus\Lambda} (1 - \pi_a(s))
\\&= 
\sum_{\Lambda\in 2^{\mathcal{A}'}, b \in \Lambda}
\prod_{a\in \Lambda} \pi_a(s)
\prod_{a\in \mathcal{A}'\setminus\Lambda} (1 - \pi_a(s))
+
\sum_{\Lambda\in 2^{\mathcal{A}'}, b \not\in \Lambda}
\prod_{a\in \Lambda} \pi_a(s)
\prod_{a\in \mathcal{A}'\setminus\Lambda} (1 - \pi_a(s))
\\&= 
\pi_b(s)
\left[
\sum_{\Lambda\in 2^{\mathcal{A}}}
\prod_{a\in \Lambda} \pi_a(s)
\prod_{a\in\mathcal{A}\setminus\Lambda} (1 - \pi_a(s))
\right]
+
(1-\pi_b(s))
\left[
\sum_{\Lambda\in 2^{\mathcal{A}}}
\prod_{a\in \Lambda} \pi_a(s)
\prod_{a\in\mathcal{A}\setminus\Lambda} (1 - \pi_a(s))
\right]
\\&= 
\pi_b(s)
\sum_{\Lambda\in 2^{\mathcal{A}}}
\pi(s,\Lambda)
+
(1-\pi_b(s))
\sum_{\Lambda\in 2^{\mathcal{A}}}
\pi(s,\Lambda)
\\&= 
\pi_b(s)
+
(1-\pi_b(s))
\\&= 1.
\end{align*}

\subsection{Proof of Proposition 2}
Define the set $\Lambda^* \subseteq \mathcal{A}$ of multiple actions as
\begin{equation*}
\Lambda^* = \left\{ a \in  \mathcal{A}  \mid \pi_a(s) \ge \frac{1}{2} \right\}.
\end{equation*}
Given a set $\Lambda \in 2^\mathcal{A}$, we use the notation $\widebar{\Lambda}$ to denote $\mathcal{A}\setminus\Lambda$.
By the definition of $\Lambda^*$, we know that
\begin{align*}
&\forall a \in \Lambda^*\ldotp \left[\pi_a(s) \ge \frac{1}{2} \ge 1-\pi_a(s)\right], \text{ and }
\\
&\forall a \in \widebar{\Lambda^*}\ldotp \left[ \pi_a(s) < \frac{1}{2} < 1-\pi_a(s)\right].
\end{align*}
Considering a set $\Lambda \in 2^\mathcal{A}$, we show that
\begin{align*}
\pi(s,\Lambda^*)
&= 
\prod_{a\in \Lambda^*} \pi_a(s)
\prod_{a\in \widebar{\Lambda^*}} (1 - \pi_a(s))
\\&= 
\prod_{a\in \Lambda^* \cap \Lambda} \pi_a(s)
\prod_{a\in \Lambda^* \cap \widebar{\Lambda}} \pi_a(s)
\prod_{a\in \widebar{\Lambda^*} \cap \Lambda} (1 - \pi_a(s))
\prod_{a\in \widebar{\Lambda^*} \cap \widebar{\Lambda}} (1 - \pi_a(s))
\\&\ge
\prod_{a\in \Lambda^* \cap \Lambda} \pi_a(s)
\prod_{a\in \Lambda^* \cap \widebar{\Lambda}} (1 - \pi_a(s))
\prod_{a\in \widebar{\Lambda^*} \cap \Lambda} \pi_a(s)
\prod_{a\in \widebar{\Lambda^*} \cap \widebar{\Lambda}} (1 - \pi_a(s))
\\&= 
\prod_{a\in \Lambda^* \cap \Lambda} \pi_a(s)
\prod_{a\in \widebar{\Lambda^*} \cap \Lambda} \pi_a(s)
\prod_{a\in \Lambda^* \cap \widebar{\Lambda}} (1 - \pi_a(s))
\prod_{a\in \widebar{\Lambda^*} \cap \widebar{\Lambda}} (1 - \pi_a(s))
\\&= 
\prod_{a\in \Lambda} \pi_a(s)
\prod_{a\in \widebar{\Lambda}} (1 - \pi_a(s))
\\&= 
\pi(s,\Lambda).
\end{align*}
Since $\pi(s,\Lambda^*) \ge \pi(s,\Lambda), \text{ for all } \Lambda \in 2^\mathcal{A}$, we have $\Lambda^* \in \argmax\limits_{\Lambda\in 2^{\mathcal{A}}} \pi(s, \Lambda)$.

\end{document}